\documentclass[conference,a4paper]{IEEEtran}

\usepackage{graphicx}
\usepackage{amsmath}
\usepackage{amssymb}
\usepackage{algorithmic}
\usepackage{algorithm}
\usepackage{subfigure}
\usepackage{array}
\usepackage{tabularx}
\usepackage{multirow}

\newcommand{\bfx}{{\textbf{x}}}

\newcommand{\bfw}{{\textbf{w}}}

\begin{document}

\title{Semi-supervised structured output prediction by local linear regression and sub-gradient descent}

\author{
\IEEEauthorblockN{Ru-Ze Liang}
\IEEEauthorblockA{
King Abdullah University of\\
Science and Technology, \\
Saudi Arabia\\
ruzeliang@outlook.com}
\and
\IEEEauthorblockN{Wei Xie}
\IEEEauthorblockA{
Vanderbilt University,\\
Nashville, TN 37235, United States\\
wei.xie@vanderbilt.edu}
\and
\IEEEauthorblockN{Weizhi Li}
\IEEEauthorblockA{
Suning Commerce R\&D Center USA, Inc\\
Palo Alto, CA 94304, United States\\
weizhili2014@gmail.com}
\and
\IEEEauthorblockN{Xin Du}
\IEEEauthorblockA{
School of Electrical Engineering,\\
Beijing Jiaotong University, \\
Beijing 100044, China}
\and
\IEEEauthorblockN{Jim Jing-Yan Wang}
\IEEEauthorblockA{
New York University Abu Dhabi,\\
United Arab Emirates}
\and
\IEEEauthorblockN{Jingbin Wang}
\IEEEauthorblockA{
National Time Service Center,\\
Chinese Academy of Sciences,\\
Xi'an 710600 , China\\
Graduate University of \\
Chinese Academy of Sciences,\\
Beijing 100039, China}
}

\maketitle

\begin{abstract}
We propose a novel semi-supervised structured output prediction method based on local linear regression in this paper. The existing semi-supervise structured output prediction methods learn a global predictor for all the data points in a data set, which ignores the differences of local distributions of the data set, and the effects to the structured output prediction. To solve this problem, we propose to learn the missing structured outputs and local predictors for neighborhoods of different data points jointly. Using the local linear regression strategy, in the neighborhood of each data point, we propose to learn a local linear predictor by minimizing both the complexity of the predictor and the upper bound of the structured prediction loss. The minimization problem is solved by sub-gradient descent algorithms. We conduct experiments over two benchmark data sets, and the results show the advantages of the proposed method.
\end{abstract}

\IEEEpeerreviewmaketitle

\section{Introduction}
\label{sec:Introduction}

Traditional machine learning methods learn model to predict binary class labels, or a continuous response \cite{wang2014effective,wang2015multiple,wang2015supervised,liu2015supervised,lin2016multi,shen2014doubly,shen2015transaction,shen2015portfolio}. However, in many machine learning applications, the forms of outputs of the prediction are structured. The structured outputs include vector, tree nodes, graph, sequence, etc. For example, in the part-of-speech tagging problem of natural language processing, given a sequence of words, we want to predict the tags of the part-of-speech of the works, and the output of the prediction is a sequence of parts-of-speech \cite{Vincze201544,Fonseca2015}. In real-world application, many outputs are not available for the training data points, leading to the problem of semi-supervised structured output prediction. Many works have been done for the problem of semi-supervised structured output prediction. Altun et al. \cite{altun2005maximum} proposed the problem of semi-supervised learning with structured outputs. Brefeld and Scheffer \cite{brefeld2006semi} proposed to solve the problem of semi-supervised structured output prediction by learning in the space of input-output space, and using co-training method. Suzuki et al. \cite{suzuki2007semi} proposed a hybrid method to solve the problem of semi-supervised structured output learning.  Jiang et al. \cite{jiang2015manifold} proposed to regularize the structured outputs by the manifold constructed from the input space directly.

All the above semi-supervised structured output prediction methods learn one single predictor for the entire data set, ignoring the different local distributions of different neighborhoods of data points. However, from our observation, the local distributions of different neighborhoods of the data may be significantly different, and this might have a significant effect to the prediction of structured outputs. Thus using one single global predictor for all the different neighborhoods are not suitable. In this paper, we learn multiple local linear structured output predictor for different neighborhoods to model the local distributions, instead of learning one single predictor for the entire data \cite{zhang2012semi,xue2009local}, and we also propose to learn the missing structured outputs for a semi- supervised data set simultaneously. For each data point,  the local distribution around this data point by its k nearest neighborhood was presented, and we try to model it by learning a local linear structured output predictor. In order to create a local linear structured output predictor, we minimize an upper bound of the structured losses of the data points in this neighborhood, as well as the squared $\ell_2$ norm of the predictor parameter vector. Some data points are shared by different neighborhoods, playing the role of bridging different local distributions. To solve the problem, we develop an iterative algorithm by using the gradient descent method.

%In this paper, we propose to learn multiple local linear structured output predictor for different neighborhoods to model the local distributions, instead of learning one single predictor for the entire data \cite{zhang2012semi,xue2009local}. Moreover, we also propose to learn the missing structured outputs for a semi-supervised data set simultaneously. For each data point, we propose to present the local distribution around this data point by its $k$ nearest neighborhood, and model it by learning a local linear structured output predictor. To learn the parameters of this local predictor, we propose to minimize a upper bound of the structured losses of the data points in this neighborhood, and the squared $\ell_2$ norm of the predictor parameter vector. Some data points are shared by different neighborhoods, and they play the role of bridging different local distributions. To solve the problem, we develop an iterative algorithm, by using gradient descent method.

The rest parts of this paper are organized as follows. In section \ref{sec:Problem},  the proposed maximum top precision similarity learning method is introduced. In section \ref{sec:Exp}, the experiments in four benchmark data sets are shown. In section \ref{sec:Conclu}, the paper is concluded with future works.

\section{Proposed method}
\label{sec:Problem}

Suppose we have a training set $\mathcal{X} = \mathcal{L}\cup \mathcal{U}$, where $\mathcal{L} =\{(\bfx_i,\overline{y}_i)\}_{i=1}^l$ contains $l$ labeled data points, $\mathcal{U} = \{\bfx_i\}_{i=l+1}^n$ contains $u = n-l$ unlabeled data points, and $\bfx_i$ and $\overline{y}_i\in \mathcal{Y}$ are the input vector and structured output of the $i$-th data point respectively. We propose to learn a local predictor for the neighborhood of each data points, and present the neighborhood of the $i$-th data point as the set of its $k$ nearest neighbors, $\mathcal{N}_i$. Given a candidate structured output, $y$, and an input feature vector, $\bfx_j\in \mathcal{N}_i$, we use a joint representation $\Phi(\bfx_j,y) \in \mathbb{R}^m$ to match them, and then we use a local linear predictor to predict the structured outputs for the data points in $\mathcal{N}_i$,

\begin{equation}
\label{equ:llr}
\begin{aligned}
y^*_j = \underset{y\in \mathcal{Y}}{\arg\max}~ \bfw_i^\top \Phi(\bfx_j,y), ~\forall~j:\bfx_j\in \mathcal{N}_i.
\end{aligned}
\end{equation}
We propose to learn the complete outputs simultaneously, $\{y_i\}|_{i=1}^n$, for all the data points. We also assume that the outputs of the data points in $\mathcal{L}$ is equal to the true outputs, i.e., $y_i = \overline{y}_i$, for $i:(\bfx_i,\overline{y}_i) \in \mathcal{L}$.

In each neighborhood, $\mathcal{N}_i$, we learn $\bfw_i$ and $y_j|_{j:\bfx_j\in \mathcal{N}_i}$ jointly by minimizing a structured loss function $\Delta$ and the squared $\ell_2$ norm of $\bfw_i$ simultaneously

\begin{equation}
\label{equ:localproblem}
\begin{aligned}
\min_{\bfw_i, y_j|_{j:\bfx_j\in \mathcal{N}_i}}~ &
\frac{1}{k} \sum_{j:\bfx_j\in \mathcal{N}_i} \Delta(y_j, y_j^*) + \frac{C}{2} \|\bfw_i\|_2^2,\\
s.t.~&
y_j = \overline{y}_j, j:(\bfx_j,\overline{y}_i)\in \mathcal{L},
\end{aligned}
\end{equation}
where $C$ is a scale parameter. The upper bound of $\Delta(y_j, y_j^*)$ is given as

\begin{equation}
\label{equ:upperbound1}
\begin{aligned}
&\Delta(y_j, y_j^*) \leq \bfw_i^\top \left ( \Phi(\bfx_j,z^*_j) - \Phi(\bfx_j,y_j) \right ) + \Delta(y_j, z^*_{i,j}),\\
\end{aligned}
\end{equation}
where

\begin{equation}
\label{equ:upperbound2}
\begin{aligned}
&z^*_{i,j} = \underset{y_j'\in \mathcal{Y}}{\arg\max}
\left [\bfw_i^\top \left ( \Phi(\bfx_j,y'_j) - \Phi(\bfx_j,y_j) \right ) + \Delta(y_j, y_j')\right ].
\end{aligned}
\end{equation}
We approximate the upper bound of the structured loss based on the lower bound approximation method of the structure learning of the Bayesian network \cite{fan2014tightening,fan2015improved}.  Fan et al. \cite{fan2014tightening} proposed to tighten the upper and lower bounds of the breadth-first branch and bound algorithm for the learning of Bayesian network structures. The informed variable groupings are used to create the pattern databases to tighten the lower bound, while the anytime learning algorithm is used to tighten the upper bound. These strategies show good performance in the learning process of the Bayesian network structures. The work of \cite{fan2014tightening} is a contribution of major significance to the bound approximation community, and our upper bound approximation method is also based on these strategies. Fan et al. \cite{fan2015improved} further proposed to improve the lower bound function of static k-cycle conflict heuristic for the learning of Bayesian network structures. This work is used to guild the search of the most promising search spaces. It uses a partition of the random variables of a data set, and  their research methodologies are based on the importance of the partition. A new partition method was proposed, and it uses the information extracted from the potentially optimal parent sets.

The minimization problem can be transferred to be the following problem,

\begin{equation}
\label{equ:localproblem1}
\begin{aligned}
\min_{\bfw_i, y_j|_{j:\bfx_j\in \mathcal{N}_i}}~ &
\frac{1}{k} \sum_{j:\bfx_j\in \mathcal{N}_i}
\left [\bfw_i^\top \left ( \Phi(\bfx_j,z^*_{i,j}) - \Phi(\bfx_j,y_j) \right )\right.\\
&\left. + \Delta(y_j, z^*_{i,j})\right ]
+ \frac{C}{2} \|\bfw_i\|_2^2\\
s.t.~&
y_j = \overline{y}_j,j:(\bfx_j,\overline{y}_i)\in \mathcal{L}.
\end{aligned}
\end{equation}
We propose to combine them into one single problem over the entire data set,

\begin{equation}
\label{equ:localproblem2}
\begin{aligned}
\min_{(\bfw_i, y_i)|_{i=1}^n}~ &\sum_{i=1}^n\left ( \frac{1}{k} \sum_{j:\bfx_j\in \mathcal{N}_i}
\left [\bfw_i^\top \left ( \Phi(\bfx_j,z^*_{i,j}) - \Phi(\bfx_j,y_j) \right ) \right. \right.\\
&\left. + \Delta(y_j, z^*_{i,j})\right ]
+\left. \frac{C}{2} \|\bfw_i\|_2^2\right )\\
s.t.~&
y_i = \overline{y}_i,~i:(\bfx_i,\overline{y}_i)\in \mathcal{L}.
\end{aligned}
\end{equation}
To solve the problem in (\ref{equ:localproblem2}), we use an iterative algorithm. In each iteration, we first update $\bfw_i|_{i=1}^n$, and then update $y_i|_{i=1}^n$ one by one.

\begin{itemize}
\item \textbf{Updating $\bfw_i$ by sub-gradient descent algorithm} If we only consider $\bfw_i|_{i=1}^n$, the problem in (\ref{equ:localproblem2}) is transferred to

\begin{equation}
\label{equ:localproblem3}
\begin{aligned}
\min_{\bfw_i|_{i=1}^n}~ &\sum_{i=1}^n\left ( \frac{1}{k} \sum_{j:\bfx_j\in \mathcal{N}_i}
\left [\bfw_i^\top \left ( \Phi(\bfx_j,z^*_{i,j}) - \Phi(\bfx_j,y_j) \right )\right ] \right. \\
&\left.
\vphantom{
\left \{ \frac{1}{k} \sum_{j:\bfx_j\in \mathcal{N}_i}
\left [\bfw_i^\top \left ( \Phi(\bfx_j,z^*_{i,j}) - \Phi(\bfx_j,y_j) \right ) + \Delta(y_j, z^*_{i,j})\right ] \right.
}
+ \frac{C}{2} \|\bfw_i\|_2^2 = g(\bfw_i)\right ),
\end{aligned}
\end{equation}
We use the sub-gradient descent algorithm to update $\bfw_i$, and the sub-gradient function of $g(\bfw_i)$ is

\begin{equation}
\label{equ:localproblem4}
\begin{aligned}
\nabla g(\bfw_i) =  \frac{1}{k} \sum_{j:\bfx_j\in \mathcal{N}_i}
\left [ \left ( \Phi(\bfx_j,z^*_{i,j}) - \Phi(\bfx_j,y_j) \right ) \right ]
+ C\bfw_i,
\end{aligned}
\end{equation}
and $\bfw_i$ is updated as follows,

\begin{equation}
\label{equ:updatewi}
\begin{aligned}
\bfw_i &\leftarrow \bfw_i - \eta \nabla g(\bfw_i).
\end{aligned}
\end{equation}

\item \textbf{Updating $y_i|_{i=1}^n$} If we only consider $y_i|_{i=1}^n$, the problem of (\ref{equ:localproblem2}) is transferred to

\begin{equation}
\label{equ:localproblem5}
\begin{aligned}
\min_{y_i|_{i=1}^n}~ &\sum_{i=1}^n\left ( \frac{1}{k} \sum_{j:\bfx_j\in \mathcal{N}_i}
\left [\Delta(y_j, z^*_{i,j}) - \bfw_i^\top \Phi(\bfx_j,y_j) \right ] \right )\\
s.t.~&
y_i = \overline{y}_i,~i:(\bfx_j,\overline{y}_i)\in \mathcal{L}.
\end{aligned}
\end{equation}
We solve the $n$ outputs one by one, and the problem in (\ref{equ:localproblem5}) is reduced to be

\begin{equation}
\label{equ:localproblem6}
\begin{aligned}
\min_{y_i}~ &\sum_{i':\bfx_i \in \mathcal{N}_{i'}}\left \{ \frac{1}{k} \left [\Delta(y_i, z^*_{i',i}) - \bfw_{i'}^\top \Phi(\bfx_i,y_i) \right ] \right \}\\
s.t.~&
y_i = \overline{y}_i,if~(\bfx_i, \overline{y}_i)\in \mathcal{L},
\end{aligned}
\end{equation}
with regard to only one output. We discuss the solution of this problem in two cases.

\begin{itemize}
\item \textbf{$(\bfx_i, \overline{y}_i)\in \mathcal{L}$}:

\begin{equation}
\label{equ:y1}
\begin{aligned}
y_i = \overline{y}_i.
\end{aligned}
\end{equation}

\item \textbf{$\bfx_i \in \mathcal{U}$}:

\begin{equation}
\label{equ:y2}
\begin{aligned}
y_i = \underset{y\in \mathcal{Y}}{\arg\min} ~ &\sum_{i':\bfx_i \in \mathcal{N}_{i'}}\left ( \frac{1}{k} \left [\Delta(y, z^*_{i',i})\right.\right.\\
&\left.\left. - \bfw_{i'}^\top \Phi(\bfx_i,y) \right ]
\vphantom{\frac{1}{2}}
\right ).
\end{aligned}
\end{equation}

\end{itemize}

\end{itemize}

We develop an iterative algorithm to learn the local structured output predictor and the outputs jointly, as given in Algorithm 1.

\begin{itemize}
\item \textbf{Algorithm 1}. Iterative training algorithm of semi-supervised learning of local structured output predictor.
\item \textbf{Inputs}: Training set, $\mathcal{X}$.
\item {\textbf{Inputs}: Maximum iteration number, $T$}.

\item \textbf{Initialize} $(\bfw_i,y_i)|_{i=1}^n$;

\item \textbf{For $t=1,\cdots,T$}

\begin{itemize}
\item Update the upper bound parameters

\item  For $i=1,\cdots,n$

\begin{itemize}
\item For $j:\bfx_j\in \mathcal{N}_i$

\item ~~~~Update $z^*_{i,j}$ according to (\ref{equ:upperbound2})

\end{itemize}

\item Update the local predictor parameters

\item For $i=1,\cdots, n$

\item ~~~~Update $\bfw_i$ according to (\ref{equ:updatewi})

\item Update the structured outputs

\item For $i = 1,\cdots,n$

\item ~~~~Update $y_i$ according to (\ref{equ:y1}) and (\ref{equ:y2})

\end{itemize}

\item \textbf{Output}: $\bfw_i|_{i=1}^n$.

\end{itemize}

\section{Experiments}
\label{sec:Exp}

\subsection{Experiment setup}

We use two benchmark data sets in our experiments.

\begin{itemize}

\item \textbf{SUN data set}:The class labels of this image data set  is organized as a tree structure.  The tree has 15 different leaves \cite{xiao2010sun}. For each class, we randomly select 200 images from the data set to conduct our experiments, thus there are 3,000 images in our data set in total. The structured output of a data point is a leave of the tree. We code it as a vector $\bfx$. The joint representation is given as the tensor product of $\bfx$ and $y$.

\begin{equation}
\begin{aligned}
\Phi(\bfx,y) = \bfx \otimes y.
\end{aligned}
\end{equation}
The loss function between $y$ and $y^*$, $\Delta(y,y^*)$, is defined as the height of their first common ancestor.

\item \textbf{Spanish news wire article sentence data set}: This data set contains 300 sentences, and each sentence is used as a data point in our experiment \cite{sang2002a}. The length of each sentence is 9, and the corresponding output of a sentence, $y$,  is a sequence of labels of non-name and named entities. The joint representation $\Phi(\bfx,y)$ of a sentence, $\bfx$, and a sequence of labels, $y$, is defined as the histogram of state transition, and a set of emission features. The loss function to compare $y$ and $y^*$ is defined as a 0-1 loss.

\end{itemize}

We use the 10-fold cross validation strategy to split the training and test subsets. The training set is also randomly split to be a labeled set and an unlabeled set. The average structured loss over the test set is used to evaluate the performance of the proposed algorithm.

\subsection{Experimental results}

We compare the proposed algorithm to the algorithms proposed by Altun et al. \cite{altun2005maximum}, Brefeld and Scheffer \cite{brefeld2006semi}, Suzuki et al. \cite{suzuki2007semi}, and Jiang et al. \cite{jiang2015manifold}. Our algorithm is named as semi-supervised local structured output prediction algorithm (SSLSOP). The average losses of the compared algorithms over three different data sets are given in Table \ref{tab:compare}. It is obvious that the proposed algorithm outperforms all competing algorithms significantly.

\begin{table}
\centering
\caption{Average structured losses of the compared algorithms over three data sets.}
\label{tab:compare}
\begin{tabular}{|l|c|c|c|c|}
\hline
Method &  SUN data set &  Spanish news data set \\\hline\hline
SSLSOP  & \textbf{0.628}   & \textbf{0.450} \\\hline
Jiang et al. \cite{jiang2015manifold} & 0.677 & 0.492\\\hline
Altun et al. \cite{altun2005maximum}  & 0.738 & 0.504 \\\hline
Brefeld and Scheffer \cite{brefeld2006semi}  & 0.762 & 0.511 \\\hline
Suzuki et al. \cite{suzuki2007semi}  & 0.754& 0.574\\\hline
\end{tabular}
\end{table}

\begin{figure}
\centering
\includegraphics[width=0.5\textwidth]{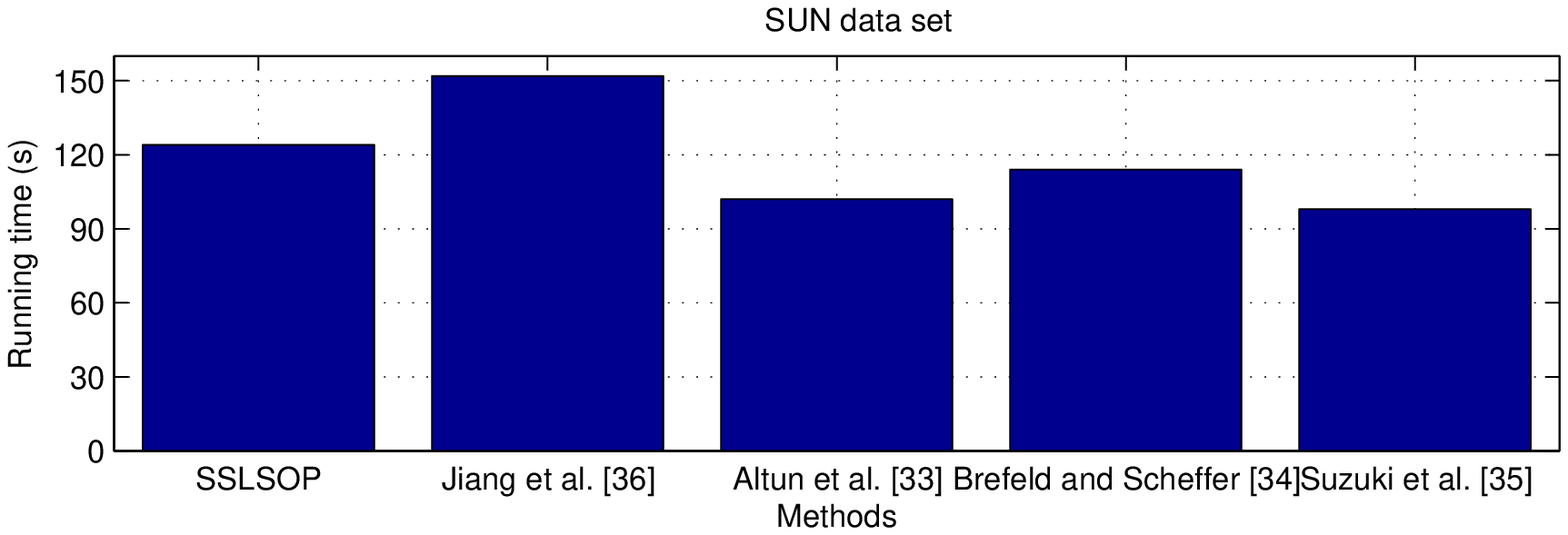}
\\
\includegraphics[width=0.5\textwidth]{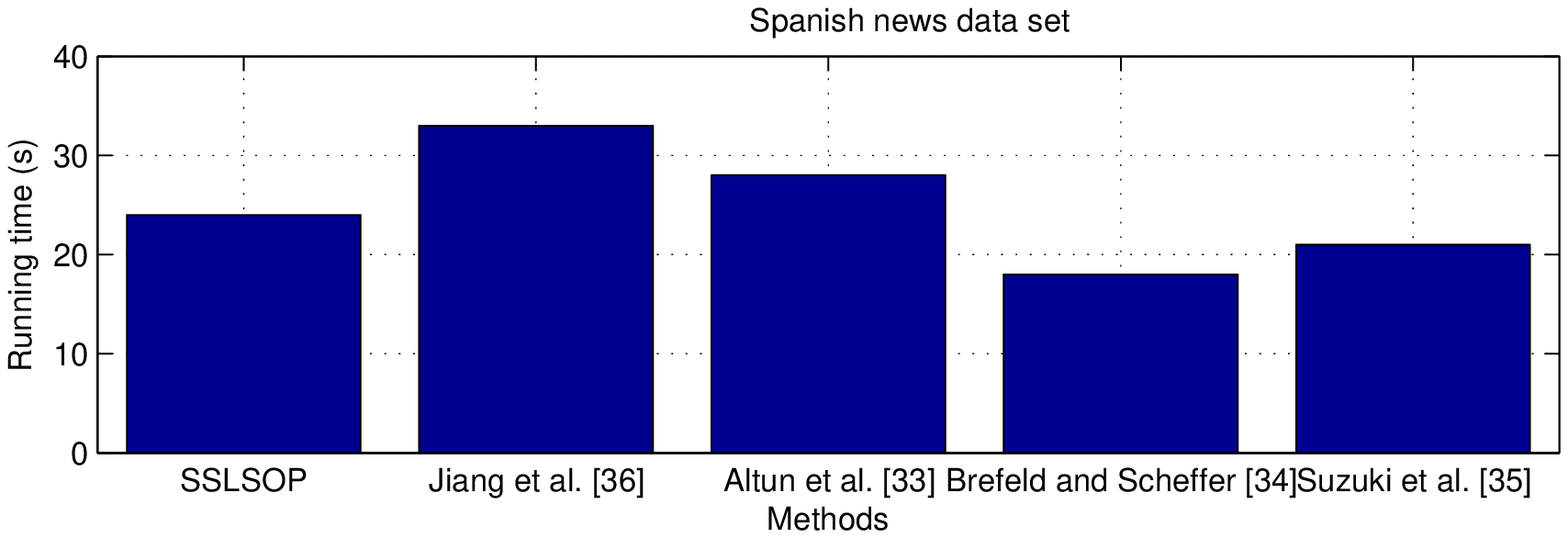}
\caption{Running time of the compared algorithms over the four data sets.}
\label{fig:Time}
\end{figure}

We are also interested in the running time of the proposed method, SSLSOP, and its competing methods. The running time of these methods over four benchmark data sets are given in Fig. \ref{fig:Time},  which shows that the proposed method, SSLSOP, consumes the second shortest running time over three data sets. The least time consuming algorithm is the one proposed by Altun et al. \cite{altun2005maximum}, however, its prediction results are not accurate.

\section{Conclusion}
\label{sec:Conclu}

In this paper, we investigate the problem of semi-supervised learning of structured output predictor. To handle the problem of diverse of the local distributions, we propose to learn local structured output predictors for neighborhoods of different data points. Moreover, we also propose to learn the missing outputs of the unlabeled data points. We build a new minimization problem to learn the local structured output predictors and the missing structured outputs simultaneously. This problem is modeled as the joint minimization of the local predictor complexity and the local structured output loss. The problem is optimized by gradient descent, and we design an iterative algorithm to learn the local predictors. The experiments are implemented over benchmark data sets including natural image classification data set and sentence part-of-speech tagging data set. In the future, we will study how to fit the proposed algorithm to big data sets, by using big data processing framework, such as Map-Reduce of Hadoop software. We also want to apply the proposed method to various applications, such as computational biology and health care \cite{wang2014computational,zhou2014biomarker,liu2013structure,peng2015modeling,xu2016mechanical,chen2015inferring,li2016supporting,xie2014securema}, computer vision \cite{manifoldlearning,generalized,xu2016image,king2015surgical,li2015outlier,thatcher2016multispectral,li2015burn,squiers2016multispectral}, natural language processing, information retrieval \cite{fan2011margin,fan2014finding,fan2010enhanced}, importance sampling \cite{shi2009new,shi2015synthesis}, and multimedia information processing \cite{xu2015semi,liang2016optimizing}.

%\bibliographystyle{IEEEtran}
%\bibliography{StructLLR}

% Generated by IEEEtran.bst, version: 1.14 (2015/08/26)

\end{document}